\documentclass[10pt,twocolumn,letterpaper]{article}

\usepackage{cvpr}              % To produce the CAMERA-READY version
\usepackage[accsupp]{axessibility}
\usepackage{array}
\usepackage{multirow}
\usepackage{amsmath}
\usepackage{graphicx}
\usepackage[dvipsnames]{xcolor}
\definecolor{cvprblue}{rgb}{0.21,0.49,0.74}
\usepackage[pagebackref,breaklinks,colorlinks,citecolor=cvprblue]{hyperref}

\def\paperID{5469} % *** Enter the Paper ID here
\def\confName{CVPR}
\def\confYear{2024}

\title{RTMO: Towards High-Performance One-Stage Real-Time Multi-Person Pose Estimation}

\author{
Peng Lu\textsuperscript{1,2},\enspace 
Tao Jiang\textsuperscript{2},\enspace
Yining Li\textsuperscript{2},\enspace
Xiangtai Li\textsuperscript{2,3},\enspace
Kai Chen\textsuperscript{2*},\enspace
Wenming Yang\textsuperscript{1}\thanks{Corresponding authors.}\\
\\
\textsuperscript{1}Tsinghua Shenzhen International Graduate School \\ \textsuperscript{2}Shanghai AI Laboratory \quad
\textsuperscript{3}Nanyang Technological University \\
{\tt\small \{lupeng, jiangtao, liyining, chenkai\}@pjlab.org.cn, yang.wenming@sz.tsinghua.edu.cn}
}

\begin{document}
\maketitle
\begin{abstract}

Real-time multi-person pose estimation presents significant challenges in balancing speed and precision. While two-stage top-down methods slow down as the number of people in the image increases, existing one-stage methods often fail to simultaneously deliver high accuracy and real-time performance. This paper introduces RTMO, a one-stage pose estimation framework that seamlessly integrates coordinate classification by representing keypoints using dual 1-D heatmaps within the YOLO architecture, achieving accuracy comparable to top-down methods while maintaining high speed. We propose a dynamic coordinate classifier and a tailored loss function for heatmap learning, specifically designed to address the incompatibilities between coordinate classification and dense prediction models. RTMO outperforms state-of-the-art one-stage pose estimators, achieving 1.1\% higher AP on COCO while operating about 9 times faster with the same backbone. Our largest model, RTMO-l, attains 74.8\% AP on COCO \texttt{val2017} and 141 FPS on a single V100 GPU, demonstrating its efficiency and accuracy. The code and models are available at \href{https://github.com/open-mmlab/mmpose/tree/main/projects/rtmo}{https://github.com/open-mmlab/mmpose/tree/main/projects/rtmo}.

\end{abstract}
    
\vspace{-0.1cm}
\section{Introduction}

\begin{figure}
\begin{centering}
\includegraphics[width=8.5cm]{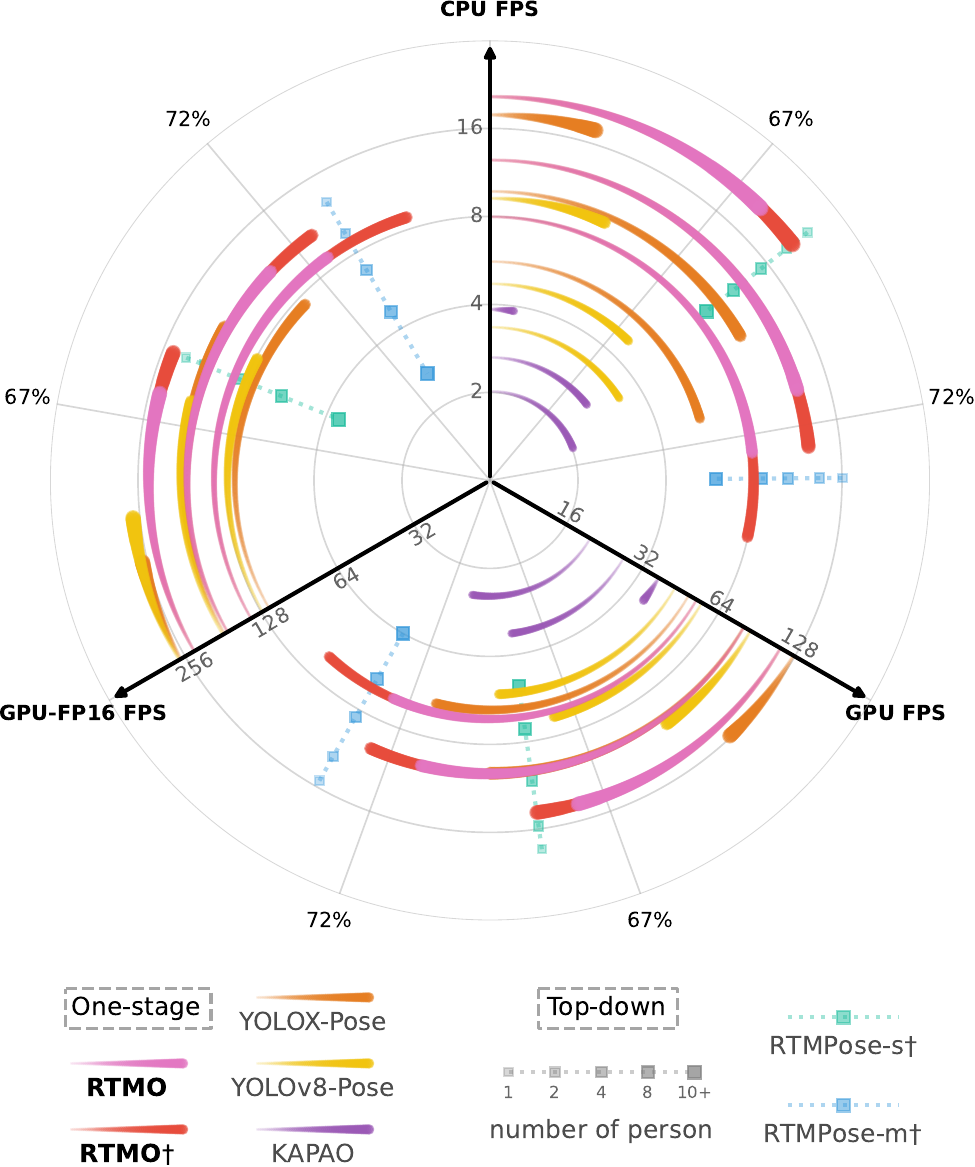}\caption{\label{fig:panel}Efficiency and efficacy comparison among real-time pose estimation methods across different inference backends and devices. The radial axes indicate inference speed in Frames Per Second (FPS). The outer circular axis shows Average Precision (AP) on the COCO \texttt{val2017} dataset. Models marked with $\dagger$ were trained with additional data beyond the COCO \texttt{train2017}.}
\vspace{-0.4cm}
\par\end{centering}
\end{figure}

Multi-person pose estimation (MPPE) is essential in the field of computer vision, with applications ranging from augmented reality to sports analytic. Real-time processing is particularly crucial for applications requiring instant feedback, such as coaching for athlete positioning. Although numerous real-time pose estimation techniques have emerged~\cite{jiang2023rtmpose,ppdet2019,maji2022yolo,yolov8}, achieving a balance between speed and accuracy remains challenging.

Current real-time pose estimation methods fall into two categories: top-down~\cite{jiang2023rtmpose,ppdet2019} and one-stage~\cite{maji2022yolo,yolov8}. Top-down methods employ pre-trained detectors to create bounding boxes around subjects, followed by pose estimation for each individual. A key limitation is that their inference time scales with the number of people in the image (see Figure~\ref{fig:panel}). On the other hand, one-stage methods directly predict the locations of keypoints for all individuals in the image. However, current real-time one-stage methods~\cite{maji2022yolo,yolov8,mcnally2022rethinking} lag in accuracy compared to top-down approaches (see Figure~\ref{fig:panel}). These methods, relying on the YOLO architecture, directly regress the keypoint coordinates, which hinders performance since this technique resembles using a Dirac delta distribution for each keypoint, neglecting the inherent ambiguity and uncertainty~\cite{gfl}. 

Alternatively, the coordinate classification methods employ dual 1-D heatmaps that increase spatial resolution by spreading the probability of keypoint locations over two sets of bins spanning the entire image. This offers more accurate predictions with minimal extra computational cost~\cite{SimCC, jiang2023rtmpose}. However, directly applying coordinate classification to dense prediction scenarios like one-stage pose estimation leads to inefficient bin utilization due to global bin distribution across the image and each person occupying only a minor region. Additionally, conventional Kullback--Leibler divergence (KLD) losses treat all samples equally, which is suboptimal for one-stage pose estimation where instance difficulty varies significantly across grids.

In this work, we overcome the above challenges and incorporate the coordinate classification approach within the YOLO-based framework, leading to the development of \textbf{\textit{R}}\textit{eal-}\textbf{\textit{T}}\textit{ime }\textbf{\textit{M}}\textit{ulti-person }\textbf{\textit{O}}\textit{ne-stage }(RTMO) pose estimation models. RTMO introduces a Dynamic Coordinate Classifier (DCC) that includes dynamic bin allocation localized to bounding boxes and learnable bin representations. Furthermore, we propose a novel loss function based on Maximum Likelihood Estimation (MLE) to effectively train the coordinate heatmaps. This new loss allows learning of per-sample uncertainty, automatically adjusting task difficulty and balancing optimization between hard and easy samples for more effective and harmonized training.

Consequently, RTMO achieves accuracy comparable to real-time top-down methods and exceeds other lightweight one-stage methods, as shown in Figure~\ref{fig:panel}. Moreover, RTMO demonstrates superior speed when processing multiple instances in an image, outpacing top-down methods with similar accuracy. Notably, RTMO-l attains a 74.8\% Average Precision (AP) on the COCO \texttt{val2017} dataset~\cite{lin2014coco} and exhibited 141 frames per second (FPS) on the NVIDIA V100 GPU. On the CrowdPose benchmark~\cite{li2019crowdpose}, RTMO-l achieves 73.2\% AP, a new state of the art for one-stage methods. The key contributions of this work include:

\begin{itemize}
% \item An innovative coordinate classification technique tailored for dense prediction scenarios, utilizing coordinate bins for precise keypoint localization while addressing the challenges posed by the varying sizes and complexities of instances.
\item An innovative coordinate classification technique tailored for dense prediction, utilizing coordinate bins for precise keypoint localization while addressing varying instance sizes and complexities.
% \item A new real-time one-stage MPPE approach that seamlessly integrates coordinate classification with the YOLO architecture, achieving an optimal balance of performance and speed among existing top-down and one-stage MPPE methods.
\item A new real-time one-stage MPPE approach that seamlessly integrates coordinate classification with the YOLO architecture, achieving an optimal balance of performance and speed among existing MPPE methods.
\end{itemize}

\section{Related Works}

\subsection{One-Stage Pose Estimator}

Inspired by advancements in one-stage object detection algorithms~\cite{tian2019fcos,liu2016ssd,yolox2021,zhou2019objects,duan2019centernet}, a series of one-stage pose estimation methods have emerged~\cite{tian2019directpose,nie2019single,zhou2019objects,dekr,maji2022yolo}. These methods perform MPPE in a single forward pass and directly regress instance-specific keypoints from predetermined root locations. Alternative approaches such as PETR~\cite{shi2022end} and ED-Pose~\cite{edpose} treat pose estimation as a set prediction problem for end-to-end keypoint regression. Beyond regression-based solutions, techniques like FCPose~\cite{mao2021fcpose}, InsPose~\cite{shi2021inspose}, and CID~\cite{cid} utilize dynamic convolution or attention mechanisms to generate instance-specific heatmaps for keypoint localization.

Compared to two-stage pose estimation methods, one-stage approaches eliminate the need for pre-processing (e.g., human detection for top-down methods) and post-processing (e.g., keypoint grouping for bottom-up methods). This results in two benefits: 1) consistent inference time, irrespective of the number of instances in the image; and 2) a simplified pipeline that facilitates deployment and practical use. Despite these advantages, the existing one-stage methods struggle to balance high accuracy with real-time inference. High-accuracy models~\cite{cid,edpose} often depend on resource-intensive backbones like HRNet~\cite{hrnet} or Swin~\cite{liu2021swin}, making real-time estimation challenging. Conversely, real-time models~\cite{mcnally2022rethinking,maji2022yolo} compromise on performance. Our model addresses this trade-off, delivering both high accuracy and fast real-time inference.

\subsection{Coordinate Classification}

SimCC~\cite{SimCC} and RTMPose~\cite{jiang2023rtmpose} have adopted coordinate classification for pose estimation, classifying keypoints into sub-pixel bins along horizontal and vertical axes to achieve spatial discrimination without high-resolution features, balancing accuracy and speed. However, spanning bins across the entire image for dense prediction methods is impractical due to the vast number of bins needed to diminish quantization error, which leads to inefficiency from many bins being superfluous for individual instances. DFL~\cite{gfl} sets bins around a predefined range near each anchor, which may not cover the keypoints of large instances and could induce significant quantization errors for small instances. Our approach assigns bins within localized regions scaled to each instance's size, which optimizes bin utilization, ensures keypoint coverage, and minimizes quantization error.

\begin{figure*}[!t]
\begin{centering}
\includegraphics[width=17.5cm]{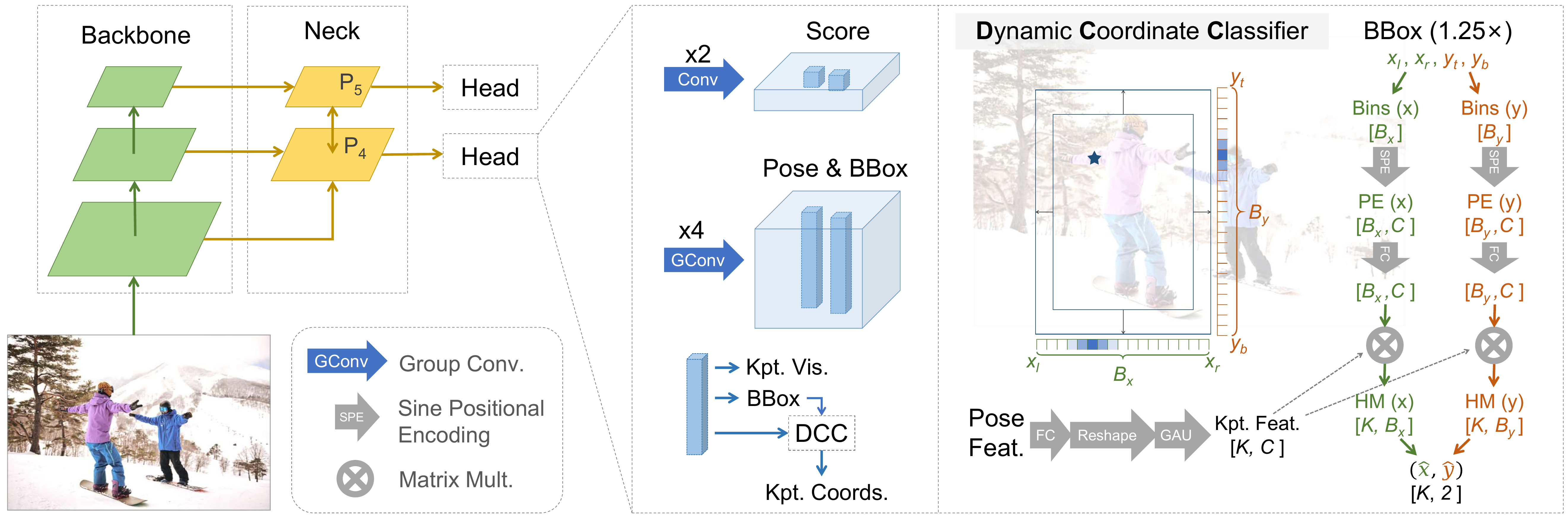}\caption{\label{fig:framework}Overview of the RTMO Network Architecture. Its head outputs predictions for the score, bounding box, keypoint coordinates and visibility for each grid cell. The Dynamic Coordinate Classifier translates pose features into \textit{K} pairs of 1-D heatmaps for both the horizontal and vertical axes, encompassing an expanded region 1.25 times the size of the predicted bounding boxes. From these heatmaps, keypoint coordinates are precisely extracted. \textit{K} denotes the total number of keypoints.
%\lxt{Please keep the font style consistent in all figures. Moreover, I think Fig. 1 and Fig. 2 can be put into one figure.}
}
\vspace{-0.3cm}
\par\end{centering}
\end{figure*}

\subsection{Transformer-Enhanced Pose Estimation}

Transformer-based architectures have become ubiquitous in pose estimation, leveraging state-of-the-art transformer backbones to improve accuracy as in  ViTPose~\cite{xu2022vitpose}, or combining transformer encoders with CNNs to capture spatial relationships~\cite{yang2021transpose}. TokenPose~\cite{li2021tokenpose} and Poseur~\cite{mao2022poseur} demonstrate the efficacy of token-based keypoint embedding in both heatmap and regression-based methods, leveraging visual cues and anatomical constraints. Frameworks like PETR~\cite{shi2022end} and ED-Pose~\cite{edpose} introduce transformers in end-to-end multi-person pose estimation, and RTMPose~\cite{jiang2023rtmpose} integrates self-attention with a SimCC-based~\cite{SimCC} framework for keypoint dependency analysis, an approach also adopted by RTMO. While positional encoding is standard in attention to inform query and key positions, we innovatively employ it to form representation vectors for each spatial bin, enabling computation of bin-keypoint similarity that facilitates accurate localization predictions.

\section{Methodology}

In our model, we adopt a YOLO-like architecture, illustrated in Figure~\ref{fig:framework}. The backbone is CSPDarknet~\cite{yolox2021}, and we process the last three feature maps using a Hybrid Encoder~\cite{lv2023detrs}, yielding spatial features $P_4$ and $P_5$ with respective downsampling rates of 16 and 32. Each pixel in these features maps to a grid cell uniformly distributed on the original image plane. The network head, utilizing dual convolution blocks at each spatial level, generates a score and corresponding pose features for every grid cell. These pose features are utilized to predict bounding boxes, keypoint coordinates, and visibility. The generation of 1-D heatmap predictions through the Dynamic Coordinate Classifier is detailed in Sec.~\ref{subsec:Dynamic-Coordinate-Classifier} while the proposed heatmap loss based on MLE is presented in Sec.~\ref{subsec:MLE}. The complete training and inference procedures are outlined in Sec.~\ref{subsec:Training-and-Inference}.

\subsection{Dynamic Coordinate Classifier\label{subsec:Dynamic-Coordinate-Classifier}}

The pose features associated with each grid cell encapsulate the keypoint displacement from the grid. Previous works~\cite{nie2019single,dekr,maji2022yolo} directly regress these displacement, and thus fall short in performance. Our study explores the integration of coordinate classification with a one-stage pose estimation framework to improve keypoint localization accuracy. A notable limitation of existing coordinate classification methods is their static strategy for bin assignment. To address this problem, we introduce the Dynamic Coordinate Classifier (DCC), which \textit{dynamically assigns ranges and forms representations for bins} in two 1-D heatmaps, effectively resolving the incompatibilities of coordinate classification in dense prediction contexts.

\paragraph{Dynamic Bin Allocation.}

Coordinate classification technique employed in top-down pose estimators allocates bins across the entire input image~\cite{SimCC,jiang2023rtmpose}, leading to bin wastage in one-stage methods since each subject occupies only a small portion. DFL~\cite{gfl} sets bins within a predefined range near each anchor, which may miss keypoints in larger instances and cause significant quantization errors in smaller ones. DCC addresses these limitations by dynamically assigning bins to align with each instance's bounding box, ensuring localized coverage. The bounding boxes are initially regressed using a pointwise convolution layer and then expanded by 1.25x to cover all keypoints, even in cases of inaccurate predictions. These expanded bounding boxes are uniformly divided into $B_{x},B_{y}$ bins along horizontal and vertical axes. The x-coordinate for each horizontal bin is calculated using:
\vspace{-0.2cm}
\begin{equation*}
x_{i}=x_{l}+\left(x_{r}-x_{l}\right)\frac{i-1}{B_{x}-1},
\end{equation*}
where $x_{r},x_{l}$ are the left and right sides of the bounding box, and index $i$ varies from 1 to $B_{x}$. The y-axis bins are computed similarly. We empirically use $B_{x}=192$ and $B_{y}=256$ for all models.

\paragraph{Dynamic Bin Encoding.}

In the context of DCC, the position of each bin varies across grids since their predicted bounding boxes differ. This differs from previous methods~\cite{SimCC, jiang2023rtmpose} where bin coordinates are fixed. Rather than a shared representation for bins across grids used in these methods, DCC generates tailored representations on-the-fly. Specifically, we encode each bin's coordinates into positional encodings to create bin-specific representations. We utilize sine positional encoding~\cite{vaswani2017attention} defined as:
\[
\left[\boldsymbol{PE}\left(x_{i}\right)\right]_{c}=\left\{ \begin{array}{lc}
\sin\left(\frac{x_{i}}{t^{c/C}}\right), & \text{for even}\ c\\
\cos\left(\frac{x_{i}}{t^{(c-1)/C}}\right), & \text{for odd}\ c
\end{array}\right.,
\]
where $t$ denotes the temperature, $c$ is the index, and $C$ represents the total number of dimension. We refine the positional encoding's adaptability for our task using a fully connected layer, which applies a learnable linear transformation $\boldsymbol{\phi}$, thereby optimizing its effectiveness in DCC.

The primary objective of DCC is to accurately predict keypoint occurrence probabilities at each bin, informed by bin coordinates and keypoint features. Keypoint features are extracted from the pose feature and refined via a Gated Attention Unit (GAU) module~\cite{Hua2022TransformerQI} following RTMPose~\cite{jiang2023rtmpose}, to enhance inter-keypoint consistency. The probability heatmap is generated by multiplying the keypoint features $\boldsymbol{f}_k$ with the positional encodings of each bin $\boldsymbol{PE}\left(x_i\right)$, followed by a softmax:
\begin{equation}
\hat{p}_{k}\left(x_{i}\right)=\frac{e^{\boldsymbol{f}_{k}\cdot\boldsymbol{\phi}\left(\boldsymbol{PE}\left({x_{i}}\right)\right)}}{\sum_{j=1}^{B_{x}}e^{\boldsymbol{f}_{k}\cdot\boldsymbol{\phi}\left(\boldsymbol{PE}\left({x_{j}}\right)\right)}},\nonumber
\end{equation}
where $\boldsymbol{f}_{k}$ is the $k$-th keypoint's feature vector.

\subsection{MLE for Coordinate Classification\label{subsec:MLE}}

In classification tasks, one-hot targets and cross-entropy loss are commonly utilized. Label smoothing, like Gaussian label smoothing used in SimCC~\cite{SimCC} and RTMPose~\cite{jiang2023rtmpose}, along with KLD, can improve performance. The Gaussian mean $\mu_x,\mu_y$ and variance $\sigma^2$ are set to the annotated coordinates and a predefined parameter. The target distribution is defined as:
\begin{equation}
p_{k}\left(x_{i} \mid \mu_x\right)=\frac{1}{\sqrt{2\pi}\sigma}e^{-\frac{\left(x_{i}-\mu_{x}\right)^{2}}{2\sigma^{2}}}\sim\mathcal{N}\left(x_{i};\mu_{x},\sigma^{2}\right).\nonumber
\end{equation}
Importantly, we note that $p_k(x_i \mid \mu_x)$ is mathematically identical to the likelihood $p_k(\mu_x \mid x_i)$ of the annotation $\mu_x$ under a Gaussian error model with true value $x_i$. This symmetrical property arises because the Gaussian distribution is symmetric with respect to its mean. Treating the predicted $\hat{p}_k(x_i)$ as the prior of $x_i$, the annotation likelihood for the $k$-th keypoint is:
\begin{align}
P\left(\mu_x\right) & =\sum_{i=1}^{B_{x}}P\left(\mu_{x} \mid x_i\right)P\left(x_i\right)\nonumber\\
 & =\sum_{i=1}^{B_{x}}\frac{1}{\sqrt{2\pi}\sigma}e^{-\frac{\left(x_{i}-\mu_{x}\right)^{2}}{2\sigma^{2}}}\hat{p}_{k}\left(x_{i}\right).\nonumber 
\end{align}
Maximizing this likelihood models the true distribution of the annotations.

In practice, we employ a Laplace distribution for $P\left(\mu_x \mid x_i\right)$ and a negative log-likelihood loss:
\begin{equation}
\mathcal{L}_{\textit{mle}}^{\left(x\right)}=-\log\left[\sum_{i=1}^{B_{x}}\frac{1}{\hat{\sigma}}e^{-\frac{\left|x_{i}-\mu_{x}\right|}{2\hat{\sigma} s}}\hat{p}_{k}\left(x_{i}\right)\right],\nonumber
\end{equation}
where the instance size $s$ normalizes the error and $\hat{\sigma}$ is the predicted variance. The constant factor is omitted as it does not affect the gradient. The total Maximum Likelihood Estimation (MLE) loss is  $\mathcal{L}_{\textit{mle}}=\mathcal{L}_{\textit{mle}}^{\left(x\right)}+\mathcal{L}_{\textit{mle}}^{\left(y\right)}$.

Unlike KLD, our MLE loss allows for learnable variance, representing uncertainty. This uncertainty learning framework automatically adjusts the difficulty of various samples~\cite{he2019bounding,chang2020data}. For hard samples, the model predicts a large variance to ease optimization. For simple samples, it predicts a small variance, aiding in accuracy improvement. With KLD, adopting a learnable variance is problematic - the model leans to predict a large variance to flatten the target distribution as this simplifies learning. More discussion can be found in Sec.~\ref{sec:ablation_study}.

\subsection{Training and Inference\label{subsec:Training-and-Inference} }

\begin{figure*}[h]
\begin{centering}
\begin{tabular}{ccc}
\includegraphics[width=5.5cm]{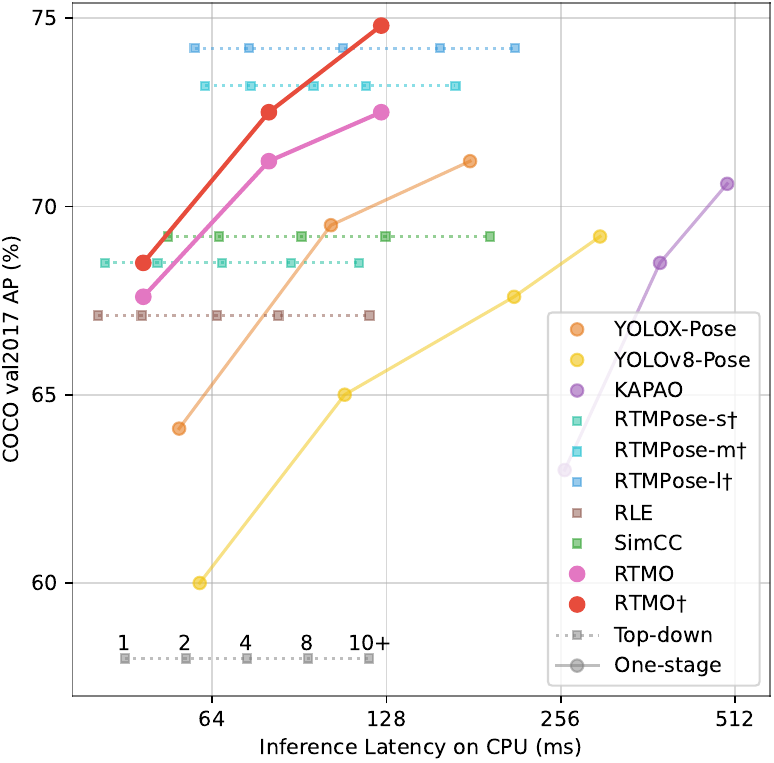} & \includegraphics[width=5.5cm]{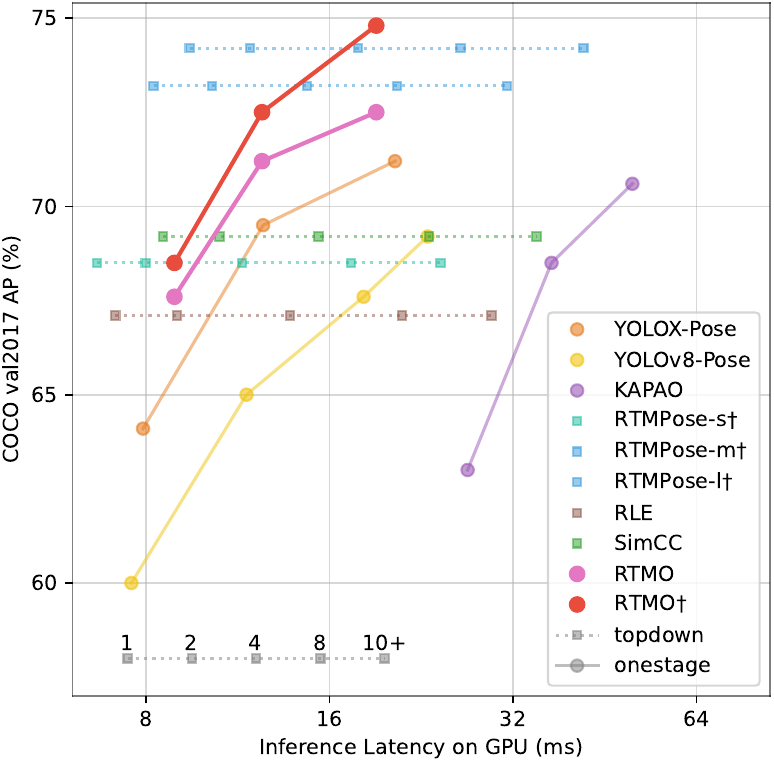} & \includegraphics[width=5.5cm]{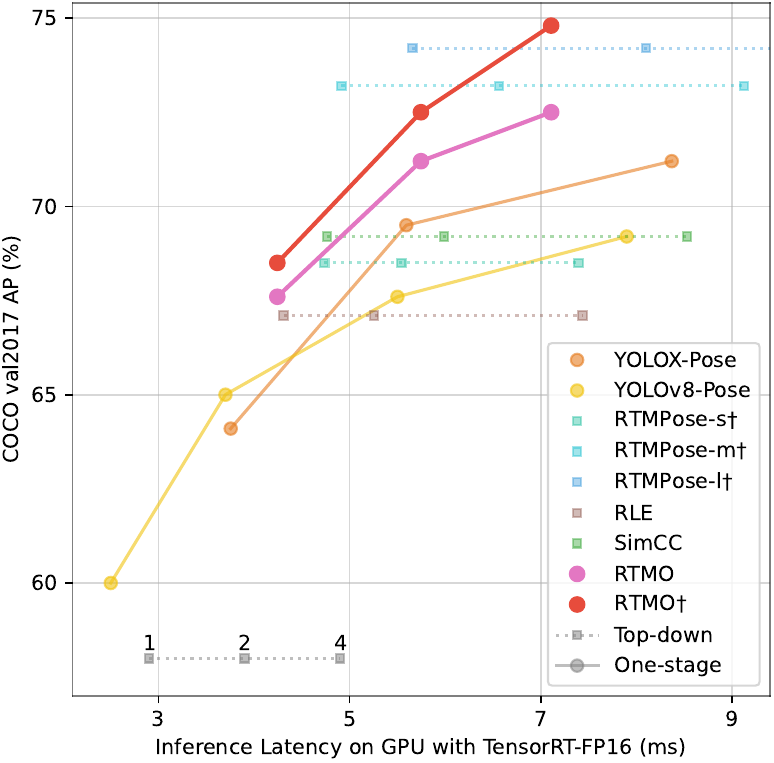}\tabularnewline
\end{tabular}\caption{\label{fig:coco_val_acc_speed}Comparison of RTMO with other real-time multi-person pose estimators. The latency for top-down methods varies depending on the number of instances in the image, as indicated by numerical values in the figures. 
All models are evaluated without test-time augmentation.
$\dagger$ indicates that the model was trained using additional data beyond the COCO \texttt{train2017} dataset.
}
\vspace{-0.3cm}
\par\end{centering}
\end{figure*}

\paragraph{Training.}

Our model, adhering to a YOLO-like structure, employs dense grid prediction for human detection and pose estimation. It is crucial for the model to differentiate between positive and negative grids. We extend SimOTA~\cite{yolox2021} for training, assigning positive grids based on grid scores, bounding box regression, and pose estimation accuracy. The head's score branch classifies these grids, supervised by varifocal loss~\cite{zhang2021varifocalnet} $\mathcal{L}_{\textit{cls}}$, with target scores being the Object Keypoint Similarity (OKS) between the predicted pose and the assigned ground truth for each grid. 

Positive grid tokens yield bounding box, keypoint coordinates, and visibility predictions. Keypoint coordinates are derived via the DCC, while other predictions come from pointwise convolution layers. The losses applied are IoU loss for bounding boxes $\mathcal{L}_{\textit{bbox}}$, MLE loss for keypoints $\mathcal{L}_{\textit{mle}}$, and BCE loss for visibility $\mathcal{L}_{\textit{vis}}$.

Given the DCC's computational demands, we implement a pointwise convolution layer for preliminary coordinate regression, similar to YOLO-Pose~\cite{maji2022yolo}, to mitigate out-of-memory issues. This regressed keypoints $\mathrm{kpt}_{\textit{reg}}$ serves as a proxy in SimOTA for positive grid selection, with the decoded keypoints $\mathrm{kpt}_{\textit{dec}}$ later used to calculate OKS. The regression branch's loss is OKS loss~\cite{maji2022yolo}:
\vspace{-0.1cm}
\begin{equation*}
\mathcal{L}_{\textit{proxy}}=1-\mathrm{OKS}\left(\mathrm{kpt}_{\textit{reg}},\mathrm{kpt}_{\textit{dec}}\right).
\end{equation*}
\vspace{-0.5cm}

The total loss for the proposed model is 
\vspace{-0.15cm}
\begin{equation}
\mathcal{L}=\lambda_{1}\mathcal{L}_{\textit{bbox}}+\lambda_{2}\mathcal{L}_{\textit{mle}}+\lambda_{3}\mathcal{L}_{\textit{proxy}}+\lambda_{4}\mathcal{L}_{\textit{cls}}+\mathcal{L}_{\textit{vis}},\nonumber
\end{equation}
with hyperparameters $\lambda_{1},\lambda_{2},\lambda_{3}$, and $\lambda_{4}$ set at $\lambda_{1}=\lambda_{2}=5, \lambda_{3}=10$, and $\lambda_{4}=2$.

\paragraph{Inference.}

During inference, our model employs a score threshold of 0.1 and non-maximum suppression for grid filtering. It then decodes pose features from selected grids into heatmaps, using integral over the heatmaps to derive keypoint coordinates. This selective decoding approach minimizes the number of features for processing, reducing computational cost. 
\section{Experiments}

\subsection{Settings}

\paragraph{Datasets.}

Experiments were primarily conducted on the COCO2017 Keypoint Detection benchmark~\cite{lin2014coco}, comprising approximately 250K person instances with 17 keypoints. Performance comparisons were made with state-of-the-art methods on both the \texttt{val2017} and \texttt{test-dev} sets. To explore our model's performance ceiling, training was also extended to include additional datasets: CrowdPose~\cite{li2019crowdpose}, AIC~\cite{JiahongWu2017AIC}, MPII~\cite{MykhayloAndriluka20142DHP}, JHMDB~\cite{jhmdb}, Halpe~\cite{alphapose}, and PoseTrack18~\cite{andriluka2018posetrack}. These annotations were converted to COCO format. RTMO was further evaluated on the CrowdPose benchmark~\cite{li2019crowdpose}, which is known for its high complexity due to crowded and occluded scenes, comprising 20K images and approximately 80K persons with 14 key points. OKS-based Average Precision (AP) served as the evaluation metric for both datasets.

\paragraph{Implementation Details. }

During training, we adopt the image augmentation pipeline from YOLOX~\cite{yolox2021}, incorporating mosaic augmentation, random color adjustments, geometric transformations, and MixUp~\cite{zhang2018mixup}. Training images are resized to dimensions {[}480, 800{]}. Epoch counts are set to 600 and 700 for the COCO and CrowdPose datasets, respectively. The training process is divided into two stages: the first involves training both the proxy branch and DCC using pose annotations, and the second shifts the target of the proxy branch to the decoded pose from DCC. The AdamW optimizer~\cite{ADAMW} is used with a weight decay of 0.05, and training is performed on Nvidia GeForce RTX 3090 GPUs with batch size 256. Initial learning rates are set to $4\times10^{-3}$ and $5\times10^{-4}$ for the two training phases, decaying to $2\times10^{-4}$ via cosine annealing. For inference, images are resized to 640. CPU latency is measured on an Intel Xeon Gold CPU using ONNXRuntime. GPU latency is tested on an NVIDIA V100 GPU with ONNXRuntime and with TensorRT using half-precision floating-point (FP16) format. MMPose~\cite{mmpose2020} toolbox is used to implement RTMO models.

\subsection{Benchmark Results}

\begin{table*}
\begin{centering}
\begin{tabular}{c|c|c|c|ccc|cc|c}
\hline 
Method & Backbone & \#Params & Time (ms) & AP & $\textrm{AP}_{50}$ & $\textrm{AP}_{75}$ & $\textrm{AP}_{M}$ & $\textrm{AP}_{L}$ & AR\tabularnewline
\hline 
DirectPose~\cite{tian2019directpose} & ResNet-50 & - & 74 & 62.2 & 86.4 & 68.2 & 56.7 & 69.8 & -\tabularnewline
DirectPose~\cite{tian2019directpose} & ResNet-101 & - & - & 63.3 & 86.7 & 69.4 & 57.8 & 71.2 & -\tabularnewline
FCPose~\cite{mao2021fcpose} & ResNet-50 & 41.7M & 68 & 64.3 & 87.3 & 71.0 & 61.6 & 70.5 & -\tabularnewline
FCPose~\cite{mao2021fcpose} & ResNet-101 & 60.5M & 93 & 65.6 & 87.9 & 72.6 & 62.1 & 72.3 & -\tabularnewline
InsPose~\cite{shi2021inspose} & ResNet-50 & 50.2M & 80 & 65.4 & 88.9 & 71.7 & 60.2 & 72.7 & -\tabularnewline
InsPose~\cite{shi2021inspose} & ResNet-101 & - & 100 & 66.3 & 89.2 & 73.0 & 61.2 & 73.9 & -\tabularnewline
CenterNet~\cite{duan2019centernet} & Hourglass & 194.9M & 160 & 63.0 & 86.8 & 69.6 & 58.9 & 70.4 & -\tabularnewline
PETR~\cite{petr} & ResNet-50 & 43.7M & 89 & 67.6 & 89.8 & 75.3 & 61.6 & 76.0 & -\tabularnewline
PETR~\cite{petr} & Swin-L & 213.8M & 133 & 70.5 & 91.5 & 78.7 & 65.2 & 78.0 & -\tabularnewline
\hline 
CID~\cite{cid} & HRNet-w32 & 29.4M & \underline{\textit{84.0}} & 68.9 & 89.9 & 76.9 & 63.2 & 77.7 & 74.6\tabularnewline
CID~\cite{cid} & HRNet-w48 & 65.4M & \underline{\textit{94.8}} & 70.7 & 90.4 & 77.9 & 66.3 & 77.8 & 76.4\tabularnewline
ED-Pose~\cite{edpose} & ResNet-50 & 50.6M & \underline{\textit{135.2}} & 69.8 & 90.2 & 77.2 & 64.3 & 77.4 & -\tabularnewline
ED-Pose~\cite{edpose} & Swin-L & 218.0M & \underline{\textit{265.6}} & 72.7 & 92.3 & 80.9 & 67.6 & 80.0 & -\tabularnewline
\hline 
KAPAO-s~\cite{mcnally2022rethinking} & CSPNet & 12.6M & \textit{26.9} & 63.8 & 88.4 & 70.4 & 58.6 & 71.7 & 71.2\tabularnewline
KAPAO-m~\cite{mcnally2022rethinking}  & CSPNet & 35.8M & \textit{37.0} & 68.8 & 90.5 & 76.5 & 64.3 & 76.0 & 76.3\tabularnewline
KAPAO-l~\cite{mcnally2022rethinking}  & CSPNet & 77.0M & \textit{50.2} & 70.3 & 91.2 & 77.8 & 66.3 & 76.8 & 77.7\tabularnewline
YOLO-Pose-s~\cite{maji2022yolo} & CSPDarknet & 10.8M & \textit{7.9} & 63.2 & 87.8 & 69.5 & 57.6 & 72.6 & 67.6\tabularnewline
YOLO-Pose-m~\cite{maji2022yolo}  & CSPDarknet & 29.3M & \textit{12.5} & 68.6 & 90.7 & 75.8 & 63.4 & 77.1 & 72.8\tabularnewline
YOLO-Pose-l~\cite{maji2022yolo}  & CSPDarknet & 61.3M & \textit{20.5} & 70.2 & 91.1 & 77.8 & 65.3 & 78.2 & 74.3\tabularnewline
\hline 
RTMO-r50 & ResNet-50 & 41.7M & \textit{15.5} & 70.9 & 91.0 & 78.2 & 65.8 & 79.1 & 75.0 \tabularnewline
RTMO-s & CSPDarknet & 9.9M & \textit{8.9} & 66.9 & 88.8 & 73.6 & 61.1 & 75.7 & 70.9\tabularnewline
RTMO-s$\dagger$ & CSPDarknet & 9.9M & \textit{8.9} & 67.7 & 89.4 & 74.5 & 61.5 & 77.2 & 71.9\tabularnewline
RTMO-m & CSPDarknet & 22.6M & \textit{12.4} & 70.1 & 90.6 & 77.1 & 65.1 & 78.1 & 74.2\tabularnewline
RTMO-m$\dagger$ & CSPDarknet & 22.6M & \textit{12.4} & 71.5 & 91.0 & 78.6 & 66.1 & 79.9 & 75.6\tabularnewline
RTMO-l & CSPDarknet & 44.8M & \textit{19.1} & 71.6 & 91.1 & 79.0 & 66.8 & 79.1 & 75.6\tabularnewline
RTMO-l$\dagger$ & CSPDarknet & 44.8M & \textit{19.1} & 73.3 & 91.9 & 80.8 & 68.3 & 81.1 & 77.4\tabularnewline
\hline 
\end{tabular}\caption{\label{tab:coco-test}Performance comparison of state-of-the-art one-stage methods on the COCO \texttt{test-dev} dataset. The symbol $\dagger$ denotes models trained with additional data beyond the COCO \texttt{train2017} dataset. Inference time in \textit{italic} are obtained using a single NVIDIA Tesla V100 GPU, while times without this emphasis are from PETR's paper~\cite{petr} and evaluated using the same device. Times underlined were measured using PyTorch due to ONNX exportation incompatibilities with those models.}
\vspace{-0.4cm}
\par\end{centering}
\end{table*}

\paragraph{COCO}

To assess RTMO against other real-time pose estimators, we measured AP and inference latency on the COCO \texttt{val2017} dataset. For one-stage methods, we considered KAPAO~\cite{mcnally2022rethinking}, YOLOv8-Pose~\cite{yolov8}, and YOLOX-Pose—an adaptation of YOLO-Pose~\cite{maji2022yolo} on YOLOX~\cite{yolox2021}. For top-down approaches, RLE~\cite{rle}, SimCC~\cite{SimCC} and RTMPose~\cite{jiang2023rtmpose} were selected for comparison. RTMDet-nano~\cite{lyu2022rtmdet}, a highly efficient object detection model, served as the human detector for top-down models. Since top-down models slow down as more people appear in the image, we partitioned the COCO \texttt{val2017} set based on person counts and assessed top-down model speeds accordingly. As shown in Fig.~\ref{fig:coco_val_acc_speed}, the RTMO series outperform comparable lightweight one-stage methods in both performance and speed. Against top-down models, RTMO-m and RTMO-l are as accurate as RTMPose-m and RTMPose-l, and faster when more people are in the image. With ONNXRuntime, RTMO matches RTMPose in speed with around four people, and with TensorRT FP16, RTMO is quicker with two or more people. This demonstrates RTMO's advantage in multi-person scenarios. Importantly, although the number of tokens processed varies with the number of people in the image, the difference in inference latency is marginal. For example, the latency of RTMO-l on a GPU in a subset with more than 10 persons is only about 0.1 ms higher than in a subset with a single person, accounting for roughly 0.5\% of the total latency.

\begin{figure*}

\begin{centering}
\includegraphics[width=17.5cm]{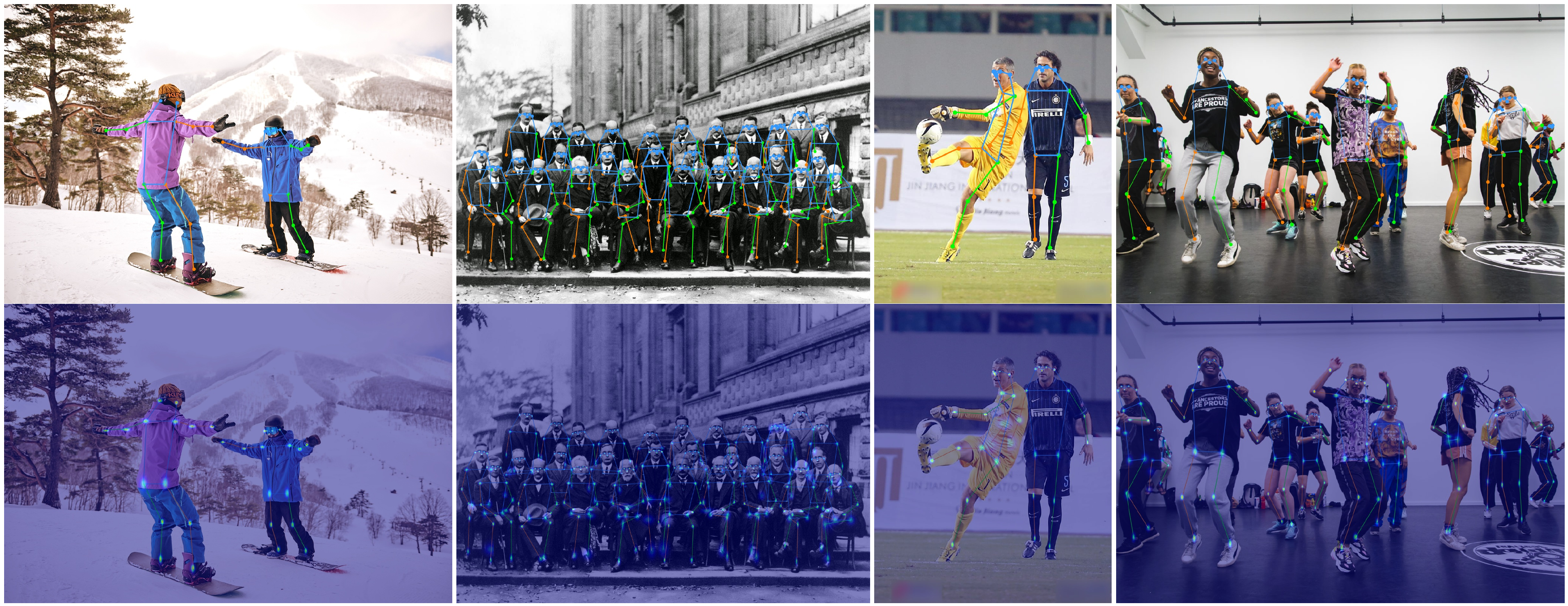}\caption{\label{fig:Visualization}Visualization of estimated human pose (top) and corresponding heatmaps (bottom).}
\par\end{centering}
\end{figure*}

Our evaluation of RTMO against leading one-stage pose estimators on the COCO \texttt{test-dev} is presented in Table~\ref{tab:coco-test}.  RTMO showcases significant advancements in speed and precision. Specifically, RTMO-s outperforms PETR~\cite{petr} using a ResNet-50~\cite{resnet} backbone, being 10x faster while maintaining similar accuracy. Compared to lightweight models like KAPAO and YOLO-Pose, RTMO consistently outperforms in accuracy across different model sizes. When trained on COCO \texttt{train2017}, RTMO-l has the second-best performance among all tested models. The highest-performing model, ED-Pose~\cite{edpose} with a Swin-L~\cite{liu2021swin} backbone, is quite heavy and not deployment-friendly. RTMO, using the same ResNet-50 backbone, surpassed ED-Pose by 1.1\% in AP and was faster. Additionally, transferring ED-Pose to the ONNX format resulted in a higher latency than its PyTorch model, about 1.5 seconds per frame. By contrast, the ONNX model of RTMO-l processes an image in just 19.1ms. With further training on additional human pose datasets, RTMO-l performs best among one-stage pose estimators in terms of accuracy.

\paragraph{CrowdPose}

\begin{table}
\begin{centering}
\begin{tabular}{>{\centering}p{2.5cm}|>{\centering}p{1.2cm}|>{\centering}p{0.5cm}>{\centering}p{0.5cm}>{\centering}p{0.5cm}>{\centering}p{0.5cm}}
\hline 
Method & \#Params & AP & $\textrm{AP}_{E}$ & $\textrm{AP}_{M}$ & $\textrm{AP}_{H}$\tabularnewline
\hline 
\multicolumn{6}{c}{Top-down methods}\tabularnewline
\hline 
{\small{}Sim.Base.~\cite{xiao2018simple}} & {\small{}34.0M} & {\small{}60.8} & {\small{}71.4} & {\small{}61.2} & {\small{}51.2}\tabularnewline
{\small{}HRNet~\cite{hrnet}} & {\small{}28.5M} & {\small{}71.3} & \textbf{\small{}80.5} & {\small{}71.4} & {\small{}62.5}\tabularnewline
{\small{}TransPose-H~\cite{yang2021transpose}} & - & {\small{}71.8} & {\small{}79.5} & {\small{}72.9} & {\small{}62.2}\tabularnewline
{\small{}HRFormer-B~\cite{yuan2021hrformer}} & {\small{}43.2M} & {\small{}72.4} & \underline{\small{}80.0} & {\small{}73.5} & {\small{}62.4}\tabularnewline
{\small{}RTMPose-m~\cite{jiang2023rtmpose}} & {\small{}13.5M} & {\small{}70.6} & {\small{}79.9} & {\small{}71.9} & {\small{}58.2}\tabularnewline
\hline 
\multicolumn{6}{c}{Bottom-up methods}\tabularnewline
\hline 
{\small{}OpenPose~\cite{openpose}} & {\small{}-} & {\small{}-} & {\small{}62.7} & {\small{}48.7} & {\small{}32.3}\tabularnewline
{\small{}HrHRNet~\cite{cheng2020higherhrnet}} & {\small{} 63.8M} & {\small{}65.9} & {\small{}73.3} & {\small{}66.5} & {\small{}57.9}\tabularnewline
{\small{}DEKR~\cite{dekr}} & {\small{}65.7M} & {\small{}67.3} & {\small{}74.6} & {\small{}68.1} & {\small{}58.7}\tabularnewline
{\small{}SWAHR~\cite{luo2021rethinking}} & {\small{} 63.8M} & {\small{}71.6} & {\small{}78.9} & {\small{}72.4} & {\small{}63.0}\tabularnewline
\hline 
\multicolumn{6}{c}{One-stage methods}\tabularnewline
\hline 
{\small{}PETR~\cite{petr}} & {\small{}220.5M} & {\small{}71.6} & {\small{}77.3} & {\small{}72.0} & \textbf{\small{}65.8}\tabularnewline
{\small{}CID~\cite{cid}} & {\small{}65.4M} & {\small{}72.3} & {\small{}78.7} & {\small{}72.1} & {\small{}64.8}\tabularnewline
{\small{}KAPAO-l~\cite{mcnally2022rethinking}} & {\small{}77.0M} & {\small{}68.9} & {\small{}76.6} & {\small{}69.9} & {\small{}59.5}\tabularnewline
{\small{}ED-Pose~\cite{edpose}} & {\small{}218.0M} & \underline{\small{}73.1} & \textbf{\small{}80.5} & \underline{\small{}73.8} & {\small{}63.8}\tabularnewline
{\small{}ED-Pose$\dagger$~\cite{edpose}} & {\small{}218.0M} & {\small{}76.6} & {\small{}83.3} & {\small{}77.3} & {\small{}68.3}\tabularnewline
\hline 
{\small{}RTMO-s} & {\small{}9.9M} & {\small{}67.3} & {\small{}73.7} & {\small{}68.2} & {\small{}59.1}\tabularnewline
{\small{}RTMO-m} & {\small{}22.6M} & {\small{}71.1} & {\small{}77.4} & {\small{}71.9} & {\small{}63.4}\tabularnewline
{\small{}RTMO-l} & {\small{}44.8M} & \textbf{\small{}73.2} & {\small{}79.2} & \textbf{\small{}74.1} & \underline{\small{}65.3}\tabularnewline
{\small{}RTMO-l$\dagger$} & {\small{}44.8M} & {\small{}83.8} & {\small{}88.8} & {\small{}84.7} & {\small{}77.2} \tabularnewline
\hline 
\end{tabular}
\caption{\label{tab:crowdpose}Performance comparisons with state-of-the-art methods on CrowdPose. The highest and second-highest performances are highlighted in bold and underlined, respectively. $\dagger$ indicates that the model was trained using additional data beyond CrowdPose.}
\vspace{-0.6cm}
\par\end{centering}
\end{table}

To evaluate RTMO under challenging scenarios, we test it on the CrowdPose~\cite{li2019crowdpose} benchmark, characterized by images with dense crowds, significant person overlap, and occlusion. The results are summarized in Table~\ref{tab:crowdpose}. Among bottom-up and single-stage approaches, RTMO-s has accuracy comparable to DEKR~\cite{dekr}, yet it uses only 15\% of the parameters. When trained on the CrowdPose dataset, RTMO-l surpasses ED-Pose~\cite{edpose} which uses a Swin-L~\cite{liu2021swin} backbone, despite having a smaller model size. Notably, RTMO-l exceeds ED-Pose primarily on medium and hard samples, demonstrating its effectiveness in challenging situations. Moreover, with additional training data, RTMO-l achieves a state-of-the-art 81.7\% AP, highlighting the model's capacity.

\subsection{Qualitative Results}

RTMO utilizes coordinate classification and demonstrates strong performance in challenging multi-person scenarios with small individuals and frequent occlusions. Figure~\ref{fig:Visualization} reveals that RTMO generates spatially accurate heatmaps even under these difficult conditions, facilitating robust and context-aware predictions for each keypoint.

\subsection{Ablation Study\label{sec:ablation_study}}

\begin{table}
\begin{centering}
    
\begin{tabular}{lccccc}
\hline 
\multirow{2}{*}{Decoding} & \multirow{2}{*}{loss} & \multicolumn{2}{c}{COCO} & \multicolumn{2}{c}{CrowdPose}\tabularnewline
\cline{3-6} \cline{4-6} \cline{5-6} \cline{6-6} 
 &  & AP & AR & AP & AR\tabularnewline
\hline 
Regression & OKS & 65.6 & 69.9 & 66.1 & 70.9\tabularnewline
CC+DBA+DBE & KLD & 64.4 & 68.2 & 62.5 & 67.5\tabularnewline
CC & MLE & 66.7 & 70.6 & 65.8 & 70.7\tabularnewline
CC+DBA & MLE & 65.6 & 70.0 & 65.2 & 70.2\tabularnewline
CC+DBA+DBE & MLE & \textbf{67.6} & \textbf{71.4} & \textbf{67.2} & \textbf{72.3}\tabularnewline
\hline 
\end{tabular}\caption{\label{tab:decode_sup}Comparison of decoding and supervision methods on COCO \texttt{val2017} and CrowdPose. The base model is RTMO-s. The term CC denotes Coordinate Classification; DBA and DBE denote Dynamic Bin Allocation and Dynamic Bin Encoding. }
\vspace{-0.4cm}
\par\end{centering}

\end{table}

\paragraph{Classification v.s. Regression. }

To assess the effectiveness of coordinate classification against regression, we replaced the model's 1-D heatmap generation with a fully connected layer for regression, supervised by the OKS loss~\cite{maji2022yolo}. Table~\ref{tab:decode_sup} compares the performances. Using the DCC module and MLE loss, coordinate classification outperforms regression with 2.0\% AP on the COCO.

\paragraph{Losses for Coordinate Classification. }

\begin{figure}
\centering{}
\vspace{-0.1cm}
\begin{tabular}{ccc}
\includegraphics[width=3cm]{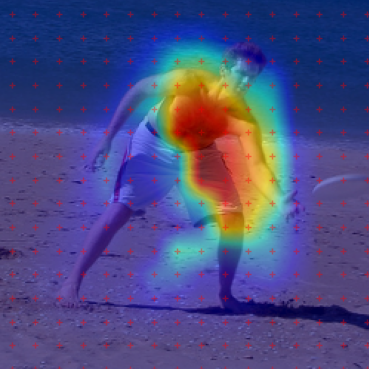} && \includegraphics[width=3cm]{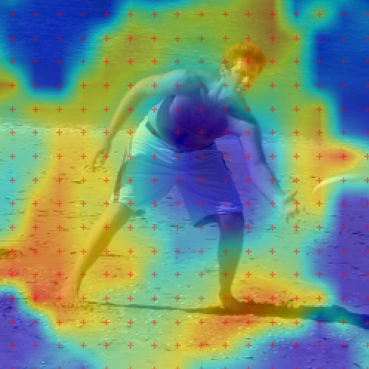}\tabularnewline
\end{tabular}
\vspace{-0.3cm}
\caption{\label{fig:Vis-var} Visualization of (left) OKS showing sample difficulty and (right) learned variance in MLE loss. Red crosses mark the position of grids.}
\end{figure}

Compare to other pose estimation methods with coordinate classification that use KLD loss, our research indicates its inadequacy for RTMO. Table~\ref{tab:decode_sup} demonstrates our MLE loss achieves higher accuracy than KLD. This improvement stems from the learnable variance in the MLE loss, which helps to balance the learning between hard and easy samples. In a one-stage pose estimator, the difficulty varies per grid due to factors like instance pose, size, and relative grid position, as visualized in Figure~\ref{fig:Vis-var}. Grids with higher OKS (easier) have lower variance in MLE loss, and vice versa. KLD fails to account for this variability, making it less effective in this context.

\paragraph{Dynamic Strategy in Coordinate Classification.}

\begin{figure}
\begin{centering}
\begin{tabular}{cc}
\includegraphics[width=3.5cm]{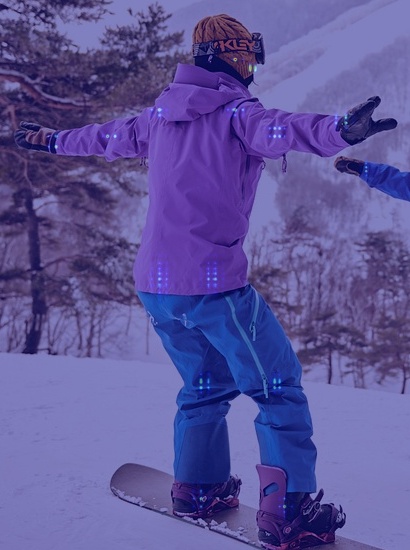} & \includegraphics[width=3.5cm]{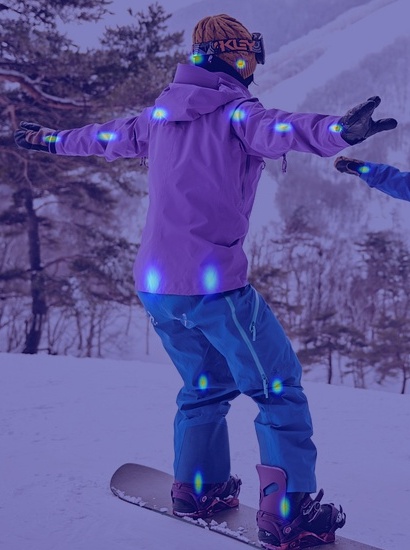}\tabularnewline
\end{tabular}
\vspace{-0.2cm}
\caption{\label{fig:heatmap_without_pe}Heatmaps decoded without (left) and with (right) DBE. }
\par\end{centering}
\vspace{-0.3cm}
\end{figure}

We firstly adopt a static coordinate classification strategy similar to DFL~\cite{gfl} where bins are distributed in a fixed range around each grid. This approach outperformed the regression method on the COCO dataset but underperformed on CrowdPose. Introducing the Dynamic Bin Allocation (DBA) strategy to this baseline resulted in decreased performance on both datasets. This is reasonable, as the semantics of each bin vary across different samples without corresponding representation adjustments. This issue was rectified by incorporating Dynamic Bin Encoding (DBE). With DBE, our DCC methods exceeded the efficacy of the static strategy on both datasets. Furthermore, Without dynamic bin encoding (DBE), the probabilities of nearby bins can vary significantly, as shown in Figure~\ref{fig:heatmap_without_pe}, contradicting the expectation that adjacent spatial locations should have similar probabilities. In contrast, incorporating DBE leads to smoother output heatmaps, indicating improved decoder training by enabling representation vectors that better capture similarities between nearby locations.

\paragraph{Feature Maps Selection.}

\begin{table}

\begin{centering}
\begin{tabular}{c|c|cc|cc}
\hline 
\multirow{2}{*}{Model} & \multirow{2}{*}{features} & \multicolumn{2}{c|}{Latency (ms)} & \multicolumn{2}{c}{Accuracy}\tabularnewline
\cline{3-6} \cline{4-6} \cline{5-6} \cline{6-6} 
 &  & CPU  & GPU  & AP & AR\tabularnewline
\hline 
\multirow{2}{*}{RTMO-s} & {\small{}\{P3, P4, P5\}} & {\small{}65.3} & {\small{}8.96} & {\small{}67.6} & {\small{}71.8}\tabularnewline
 & {\small{}\{P4, P5\}} & {\small{}48.7} & {\small{}8.91} & {\small{}67.6} & {\small{}71.4}\tabularnewline
\hline 
\multirow{2}{*}{RTMO-m} & {\small{}\{P3, P4, P5\}} & {\small{}108.6} & {\small{}16.87} & {\small{}71.4} & {\small{}75.1}\tabularnewline
 & {\small{}\{P4, P5\}} & {\small{}80.2} & {\small{}12.40} & {\small{}71.2} & {\small{}75.2}\tabularnewline
\hline 
\multirow{2}{*}{RTMO-l} & {\small{}\{P3, P4, P5\}} & {\small{}186.3} & {\small{}22.01} & {\small{}72.7} & {\small{}76.9}\tabularnewline
 & {\small{}\{P4, P5\}} & {\small{}125.4} & {\small{}19.09} & {\small{}72.5} & {\small{}76.6}\tabularnewline
\hline 
\end{tabular}\caption{\label{tab:feat_2_3}Comparison of performance and latency for models using 2 or 3 features. Accuracy metrics are based on the COCO \texttt{val2017} dataset. Latency measurements for both CPU and GPU are taken using ONNXRuntime.}
\par\end{centering}
\vspace{-0.3cm}
\end{table}

Feature pyramids~\cite{kim2018parallel} leverage multi-scale features for detecting instances of varying sizes; deeper features typically detect larger objects. Our initial model used {P3, P4, P5} features with strides of 8, 16, and 32 pixels. However, P3 contributed 78.5\% of the FLOPs in the model head while accounting for 10.7\% of correct detections. To improve efficiency, we focused on {P4, P5}. As shown in Table \ref{tab:feat_2_3}, omitting P3 led to significant speed gains with minimal accuracy loss, indicating that P4 and P5 alone are effective for multi-person pose estimation. This suggests that the role of P3 in detecting smaller instances can be compensated by the remaining features.
\section{Conclusion}

In conclusion, our RTMO model significantly improves the speed-accuracy tradeoff in one-stage multi-person pose estimation. By integrating coordinate classification within a YOLO-based framework, we achieve both real-time processing and high precision. Our approach, featuring a dynamic coordinate classifier and a loss function based on maximum likelihood estimation, effectively improves the location precision in dense prediction models. This breakthrough not only enhances pose estimation, but also establishes a robust foundation for future advancements in the scope of dense prediction for visual detection tasks.

\section*{Acknowledgments}
We thank the reviewers for their helpful comments. This work was supported by the National Key R\&D Program of China (No. 2022ZD0161600) and the Special Foundations for the Development of Strategic Emerging Industries of Shenzhen (Nos. JCYJ20200109143035495 \& CJGJZD20210408092804011)

{
    \small
    \bibliographystyle{ieeenat_fullname}
    \bibliography{main}
}


\end{document}